\documentclass[letterpaper, 10 pt, conference]{ieeeconf}  

\IEEEoverridecommandlockouts                              

\usepackage{graphics} 
\usepackage{epsfig} 
\usepackage{mathptmx} 
\usepackage{amsmath} 
\usepackage{amssymb}  
\usepackage{subfigure}
\usepackage{array}
\usepackage{placeins}
\usepackage{rotating}
\usepackage{multicol}
\usepackage{multirow}
\usepackage[table]{xcolor}
\definecolor{Gray}{gray}{0.85}
\usepackage{booktabs}
\usepackage{adjustbox}
\usepackage{chngcntr}
\usepackage{xcolor}
\usepackage{bm}
\usepackage{hyperref}

\title{\LARGE \bf
Memory Efficient Experience Replay for Streaming Learning}

\author{Tyler L. Hayes$^{\dagger}$, Nathan D. Cahill$^{\dagger}$, and Christopher Kanan$^{\dagger}$
\thanks{*This work was supported by a DARPA/ARL L2M grant [W911NF-18-2-0263] and an ONR grant [N00173-18-P-0106]. The views and conclusions contained herein are those of the authors and should not be interpreted as  representing the official policies or endorsements of any sponsor.}
\thanks{$^{\dagger}$Carlson Center for Imaging Science, Rochester Institute of Technology, Rochester, NY 14623, USA
        {\tt\small \{tlh6792,ndcsma,kanan\}@rit.edu}}
        }

\begin{document}

\maketitle
\thispagestyle{empty}
\pagestyle{empty}

\begin{abstract}
In supervised machine learning, an agent is typically trained once and then deployed. While this works well for static settings, robots often operate in changing environments and must quickly learn new things from data streams. In this paradigm, known as streaming learning, a learner is trained online, in a single pass, from a data stream that cannot be assumed to be independent and identically distributed (iid). Streaming learning will cause conventional deep neural networks (DNNs) to fail for two reasons: 1) they need multiple passes through the entire dataset; and 2) non-iid data will cause catastrophic forgetting. An old fix to both of these issues is rehearsal. To learn a new example, rehearsal mixes it with previous examples, and then this mixture is used to update the DNN. Full rehearsal is slow and memory intensive because it stores all previously observed examples, and its effectiveness for preventing catastrophic forgetting has not been studied in modern DNNs. Here, we describe the ExStream algorithm for memory efficient rehearsal and compare it to alternatives. We find that full rehearsal can eliminate catastrophic forgetting in a variety of streaming learning settings, with ExStream performing well using far less memory and computation.
\end{abstract}

\section{Introduction}
\label{intro}

Often, a robot needs to quickly learn something new. For example, a child might teach a toy robot her new friends' names and immediately test it. While deep neural networks (DNNs) are state-of-the-art for machine perception tasks, conventional models are ill-suited for this example. They learn slowly via multiple passes through a fixed training dataset, and they cannot easily be updated without suffering from catastrophic forgetting~\cite{french1999catastrophic}. We endeavor to overcome these problems to enable DNNs to be used for streaming classification\footnote{Streaming learning is sometimes also called single pass online learning.}. In streaming classification, 
\begin{enumerate}
\item New knowledge can be used immediately, 
\item Learners see each labeled instance only once, 
\item The data stream may be non-iid and structured, and
\item Learners must limit their memory usage. 
\end{enumerate}
Streaming classification is distinct from incremental batch learning, which has recently received much attention~\cite{kemker2018fearnet,kemker2018forgetting,li2016learning,lopez2017gradient,rebuffi2016icarl,zenke2017continual}, and it has unique challenges. 
Streaming classification is needed by robots that must quickly learn and make inferences in changing environments. 

\begin{figure}[t]
\begin{center} 
    \includegraphics[width=0.85\linewidth]{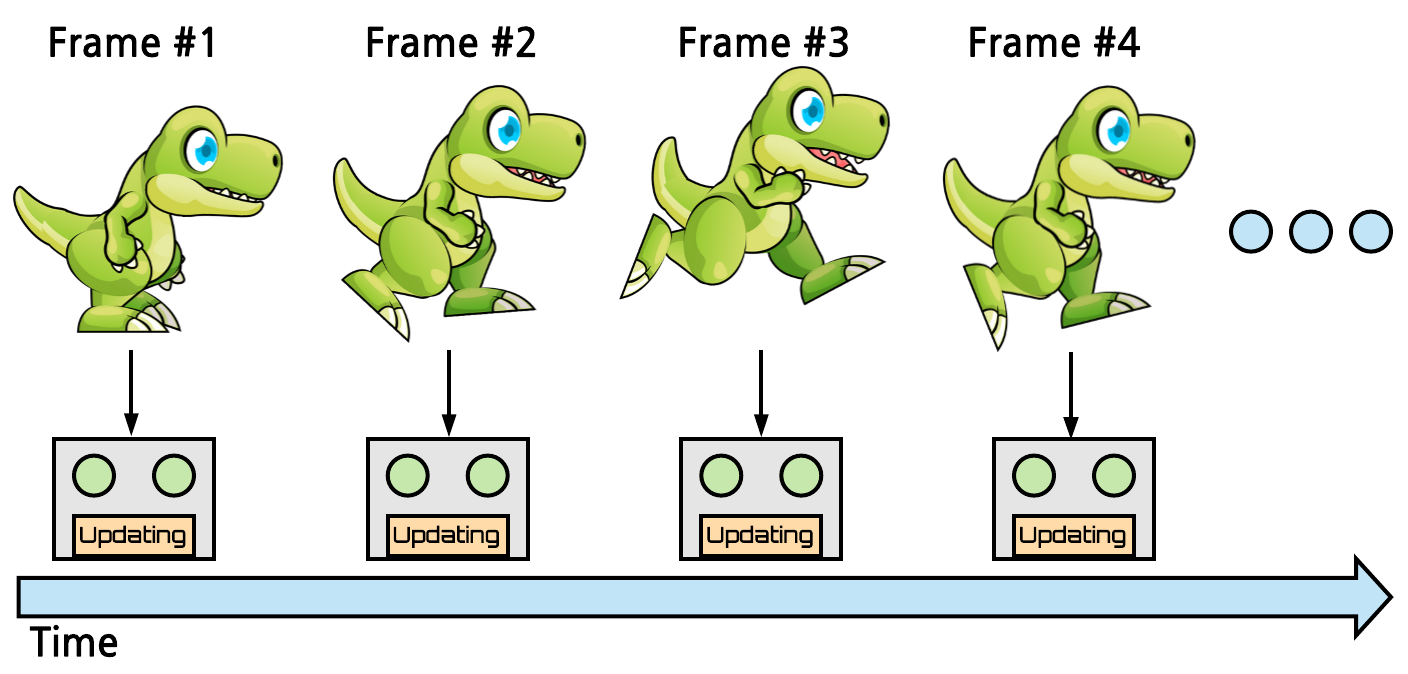}
\end{center}
  \caption{A robot may frequently encounter non-iid streams of labeled data, e.g., when learning to recognize a particular object in its environment from multiple different views, a robot would receive many temporally ordered examples of the same class. Streaming learning addresses this situation.}
\label{fig: main figure}
\end{figure}

In streaming classification, a learner receives a temporally ordered sequence of \emph{possibly} labeled input feature vectors $\left( \mathbf{x}_1 , y_1 \right), \left( \mathbf{x}_2 , y_2 \right), \ldots, \left( \mathbf{x}_T , y_T \right)$, i.e., 
$\mathcal{D} =  \left\{{\left( {{\mathbf{x}}_t ,y_t } \right)}\right\}_{t=1}^T$, where $\mathbf{x}_t \in \mathbb{R}^d$ and $y_t$ is a class label (see Fig.~\ref{fig: main figure}). Any $y_t$ that is not given must be predicted by the learner from $\mathbf{x}_t$ using the classification model built from the data observed from time $1$ to $t-1$. In this paper, we study how to do streaming classification in DNNs using memory efficient rehearsal.

In the 1980s, researchers realized updating a DNN with new information often results in catastrophic forgetting of previously learned information~\cite{mccloskey1989}. Rehearsal is one of the earliest methods for preventing this phenomenon~\cite{lin1992self}. Rehearsal, also known as replay and interleaved learning~\cite{mcclelland1995there}, limits catastrophic forgetting by mixing past experience with new data~\cite{lin1992self}. Originally, it was implemented by mixing an identical copy of \emph{all} previously seen examples with new data, and then this mixture was used to fine-tune the network with the new information. While full rehearsal has been used recently~\cite{Gepperth2016}, it has not been rigorously studied in recent DNN models. Rather than updating a DNN with a mixture of all previously observed data, we instead explore using memory limited buffers to update a DNN with a compressed set of prototypes. 

\textbf{This paper makes the following contributions}: 
\begin{itemize}
\item We rigorously examine the efficacy of rehearsal for streaming classification using benchmarks designed to induce catastrophic forgetting in DNNs.

\item We propose new metrics for streaming classification.

\item We show that full rehearsal stops catastrophic forgetting.

\item We study six distinct methods for making rehearsal memory efficient, including streaming clustering.

\item We introduce ExStream, a streaming learning framework for memory efficient rehearsal.

\end{itemize}

\section{Related Work}
\label{sec:related work}

\subsection{{Catastrophic Forgetting}}

Catastrophic forgetting is the dramatic loss of previously learned knowledge that often occurs when a DNN is incrementally trained with non-iid data~\cite{french1999catastrophic}. Catastrophic forgetting is a result of the stability-plasticity dilemma~\cite{abraham2005memory}; that is, the network must maintain a balance between plasticity to acquire new knowledge and synaptic stability to maintain previously learned information. Learning a new task requires weights to change, and can cause representations needed for other tasks to be lost. 

Besides rehearsal, there are four other major ways to mitigate catastrophic forgetting in DNNs~\cite{kemker2018forgetting}: 1) regularization, 2) ensembling, 3) sparse-coding, and 4) dual-memory models. Regularization  constrains weight updates so they do not interfere with previous learning~\cite{hinton1987using, kirkpatrick2017,li2016learning,maltoni2018continuous,zenke2017continual}. Ensembling methods train multiple models and combine their outputs to make predictions~\cite{dai2007boosting,fernando2017,polikar2001learn,ren2017life,wang2003mining}. Sparse-coding methods reduce the chance of new representations interfering with older representations~\cite{FEL}. Dual-memory models are a brain-inspired approach that use two networks~\cite{Gepperth2016,kemker2018fearnet}: one that learns fast inspired by the hippocampus and one that learns slowly inspired by the neocortex~\cite{nadel1997memory}. The fast learner is then used to train the slow learner, similar to the storage buffer in rehearsal methods.

For incremental batch learning, Kemker et al.~\cite{kemker2018forgetting} showed that there was a large gap between all of these methods and an offline baseline, but GeppNet~\cite{Gepperth2016}, a rehearsal-based method, was best\footnote{Although GeppNet uses rehearsal, it is not a traditional DNN and it is not designed for streaming learning, so we do not compare against it.}. Here, we adapt rehearsal to enable streaming classification in DNNs.

\subsection{{Incremental Batch Learning}}

In incremental batch learning, a labeled training dataset $\mathcal{D}$ is organized into $T$ distinct \emph{batches} that are possibly non-iid, i.e., $\mathcal{D}=\bigcup_{t=1}^{T}B_{t}$. The learner sequentially observes each batch consisting of $N_{t}$ labeled training pairs, i.e., $B_{t}=\left\{\left(\mathbf{x}_{i}, y_{i}\right)\right\}_{i=1}^{N_{t}}$, where $\mathbf{x}_{i} \in \mathbb{R}^{d}$ is a training example and $y_{i}$ is the associated label. During time $t$, the learner is only allowed to learn examples from $B_{t}$. This training paradigm has recently been heavily studied~\cite{fernando2017,kemker2018fearnet,li2016learning,lopez2017gradient,rebuffi2016icarl,zenke2017continual}. Streaming learning takes incremental batch learning a step further by imposing the following constraints: 1) the batch size is equal to one, i.e., $N_{t}=1$, 2) the learner is only allowed a single pass through the labeled dataset, and 3) the learner may be evaluated at any time during training.

\subsection{{Streaming Learning}}

Streaming learning differs from incremental batch learning, because it requires learning quickly from individual instances. Streaming learning has been heavily studied in unsupervised clustering, where methods can be broken up into several categories. Partitioning algorithms separate points into $j$ disjoint clusters by minimizing an objective function~\cite{ackermann2012streamkm++,farnstrom2000scalability,guha2003clustering,kohonen1995learning}. Micro-clustering algorithms first generate local clusters based on the data stream directly, then these micro-clusters are clustered themselves to generate a global clustering (macro-clusters)~\cite{aggarwal2003framework,aggarwal2004demand,fichtenberger2013bico,kranen2011clustree,tennant2017scalable,zhang1997birch}. Density-based algorithms cluster data that lie in high-density regions of feature space together, while labeling points in low-density regions as outliers~\cite{birant2007st,cao2006density,chen2007density,hahsler2016clustering,kriegel2003incremental}. Since many data streams are high-dimensional, there is also work focused on projected subspace clustering~\cite{aggarwal2004framework,chairukwattana2014se,hassani2012density,ntoutsi2012density}. Here, we use stream clustering to maintain rehearsal buffers for streaming classification with DNNs. 

While almost no work has been done on streaming classification with DNNs, there are streaming classifiers.  Many  are based on Hoeffding Decision Trees~\cite{bifet2009adaptive,domingos2000mining,gama2006decision,hulten2001mining,ikonomovska2011learning,ikonomovska2011speeding,jin2003efficient,manapragada2018extremely,read2012scalable} or ensemble methods~\cite{bifet2009new,brzezinski2014reacting,gomes2017adaptive,kolter2007dynamic,krawczyk2017ensemble,littlestone1994weighted,mahdi2018combination,pfahringer2007new,wang2003mining,wolpert1992stacked}. Both approaches are slow to train~\cite{gaber2007survey}. ARTMAP networks~\cite{carpenter2010biased,carpenter1992fuzzy,carpenter1991artmap,williamson1996gaussian} are another approach and they learn faster.  However, none of these methods learn shared representations or constrain resource usage~\cite{gaber2007survey}. For high-dimensional datasets, they all typically perform worse than offline DNNs.

\section{Memory Efficient Rehearsal}

\label{sec: buffers}

Full rehearsal, as originally proposed, limits catastrophic forgetting by mixing \emph{all} older examples with new examples to be learned. While prior work did not evaluate full rehearsal rigorously,  it is not resource efficient in terms of storage or computation. For a resource constrained robot designed to learn and operate over a long period of time, full rehearsal is not a workable solution.  However, training examples often have significant redundancy, so rather than storing all prior examples, a smaller number of prototypes that capture most of the intra-class variance could be stored instead. 

\begin{figure}[t!]
\begin{center} 
\vspace{1.7mm}
    \includegraphics[width=0.99\linewidth]{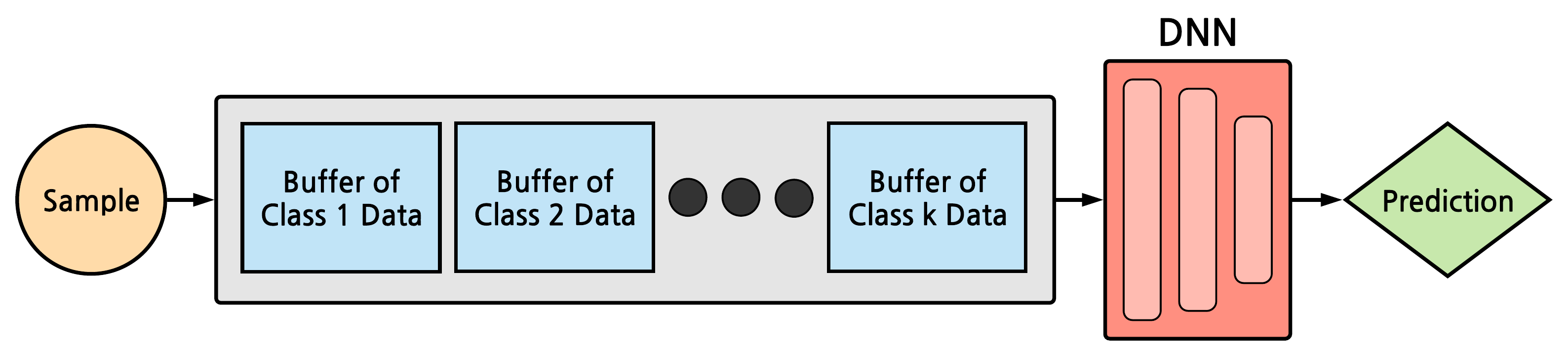}
\end{center}
   \caption{Rehearsal was developed almost 30 years ago to enable streaming learning in neural networks. In rehearsal, catastrophic forgetting is prevented by mixing previously observed examples with more recent examples during training. Here, a sample is fed into the streaming learner. The learner then adds/merges the sample with the appropriate class. All prototypes from the streaming buffers are then collected and used to update the DNN, before a final prediction is made.} 
\label{fig: rehearsal}
\end{figure}

We adapt rehearsal to $K$-way classification by having $K$ class-specific buffer data structures, where each buffer contains at most $b$ prototypes. These buffers are updated in a streaming fashion and the data stored in them is used to update (fine-tune) a DNN for classification (see Fig.~\ref{fig: rehearsal}).

When a labeled example $\left( \mathbf{x}_t , y_t \right)$ is observed, first the appropriate buffer is updated. If the buffer is not full, then $\mathbf{x}_t$ is simply copied into the buffer for class $y_t$. Otherwise, elements in this buffer must be compressed or removed to make room for encoding $\mathbf{x}_t$. After updating the buffer, data from all buffers is used to update the DNN. We update the DNN using one iteration of gradient descent for each prototype, with the order of the prototypes chosen randomly. In our experiments, after updating we evaluate the DNN on all of the test data. Since we are maintaining buffers in a streaming setting, the order of the data stream will impact the prototypes being stored, thus affecting the classification results. For this reason, we have designed several experiments, described in Sec.~\ref{sec:experiments} and Sec.~\ref{sec: results}, that evaluate how well models perform in different ordering scenarios. 

We explore six different buffers for memory limited rehearsal, and we vary the fixed capacity $b$. We use four streaming clustering methods: 1) our new Exemplar Streaming algorithm, ExStream; 2) a version of Online $k$-means~\cite{kohonen1995learning}; 3) the micro-cluster-based CluStream method~\cite{aggarwal2003framework}; and 4) the projected micro-clustering method known as HPStream~\cite{aggarwal2004framework}. We also use two replacement methods: 1) reservoir sampling~\cite{vitter1985random} and 2) a first-in, first-out queue. 

\subsection{Stream Clustering Buffers}

\subsubsection{ExStream}
We introduce a partitioning-based method for stream clustering that we call the Exemplar Streaming (ExStream) algorithm. In addition to storing clusters, ExStream also stores counts that tally the total number of points in each cluster. Once a class-specific buffer is full and a new example $\left( \mathbf{x}_{t} , y_{t} \right)$ streams in, the two closest clusters in the buffer for class $y_{t}$ are found using the Euclidean distance metric and merged together using
\begin{equation}
\mathbf{w}_{i} \leftarrow \frac{c_{i} \mathbf{w}_{i} + c_{j} \mathbf{w}_{j}}{c_{i}+c_{j}}  \enspace ,
\end{equation}
where $\mathbf{w}_{i}$ and $\mathbf{w}_{j}$ are the two closest clusters and $c_{i}$ and $c_{j}$ are their associated counts. Subsequently, the counter at $c_{i}$ is updated as the sum of the counts at locations $i$ and $j$ and the new point is inserted into the buffer at location $j$. That is, $c_{i} \leftarrow c_{i} + c_{j}$ and $\mathbf{w}_{j} \leftarrow \mathbf{x}_{t}$ with $c_{j}=1$. Source code is at: \url{https://github.com/tyler-hayes/ExStream}.

\subsubsection{Online $k$-means}
This is a partitioning-based heuristic for an online variant of the traditional $k$-means clustering algorithm~\cite{liberty2016algorithm}. This heuristic is sometimes referred to as Learning Vector Quantization~\cite{kohonen1995learning}. Similar to ExStream, this method stores a counter for each exemplar that counts how many points have been added to that exemplar/cluster. 
After the buffer is full, when a new example $\left( \mathbf{x}_{t} , y_{t} \right)$ is observed, the index $i$ of the closest exemplar to $\mathbf{x}_{t}$ in that buffer is found using the Euclidean distance metric. It is then updated by
\begin{equation}
\mathbf{w}_{i} \leftarrow \frac{c_{i} \mathbf{w}_{i} + \mathbf{x}_{t}}{c_{i} + 1}  \enspace ,
\end{equation}
where $\mathbf{w}_{i}$ is the closest stored exemplar in the buffer for class $y_{t}$ and $c_{i}$ is the counter associated with $\mathbf{w}_{i}$. Subsequently, the counter $c_{i}$ is incremented by one.

\subsubsection{CluStream}

The stream clustering approach, CluStream~\cite{aggarwal2003framework}, uses the \emph{cluster feature vector}  structure~\cite{zhang1996birch} for maintaining sets of micro-clusters online. When a point streams into the model, the distance of that point is computed to every micro-cluster. Once the closest micro-cluster is found, the \emph{maximal boundary factor} of that cluster is computed. If the distance from the point to that micro-cluster is within the boundary, the point is added to that cluster, otherwise a new cluster is created with only that point and then either the least recently updated cluster is removed, or the two closest clusters are merged. This method is not designed to handle high-dimensional data streams.

\subsubsection{HPStream}

The High-dimensional Projected Stream (HPStream) algorithm~\cite{aggarwal2004framework} is a micro-cluster-based projected clustering method for high-dimensional data. Similar to CluStream, the method uses a \emph{faded cluster structure}~\cite{aggarwal2004framework} to store statistics representative of each micro-cluster. The main advantage of the HPStream algorithm is that it maintains a \emph{bit vector} for each micro-cluster that indicates the relevant dimensions for projected clustering of the associated cluster. When a new point, $\left( \mathbf{x}_{t} , y_{t} \right)$, arrives, the algorithm first computes the bit vector for each cluster. The index of the closest micro-cluster to $\mathbf{x}_{t}$ is found and the \emph{limiting radius} of that micro-cluster is computed. If the distance from $\mathbf{x}_{t}$ to the micro-cluster is smaller than the limiting radius, the point is added to the cluster, otherwise the least recently updated cluster is replaced by the new sample.

\subsection{Replacement Buffers}
We compare the stream clustering buffers to two simple replacement baselines. For both methods, rather than compressing using clustering, they replace a stored prototype with the new input. 

\subsubsection{Reservoir Sampling} This is the traditional method for maintaining a buffer (reservoir) of random samples in an online fashion~\cite{vitter1985random}. Data from a stream of length $M$ flow into the model, sample by sample, until a class-specific buffer of size $b$ is full. After a class-specific buffer is full and a new sample from that class arrives, the new sample has a probability $b/M$ of replacing an existing sample.

\subsubsection{First-In, First-Out Queue} This is one of the most popular strategies for maintaining exemplars, and was recently used for incremental batch learning in \cite{lopez2017gradient}. In this model, data streams in until a class-specific buffer is full. Once a new sample from that class arrives, it replaces the oldest example in that class-specific buffer.

\subsection{Baselines}

We compare the memory limited streaming methods against three baselines:

\subsubsection{No Buffer} This method trains the DNN sample by sample with a single pass through the entire labeled dataset.

\subsubsection{Full Rehearsal} This method stores all training examples in an unbound buffer as they stream in and uses those examples to fine-tune the DNN.

\subsubsection{Offline DNN} This is a conventional offline DNN trained from scratch on all training data. This serves as an approximate upper bound for all experiments.

\begin{figure*}[h]
\centering
\vspace{1mm}
      \subfigure[iCub1]{
  \includegraphics[width=0.3\linewidth]{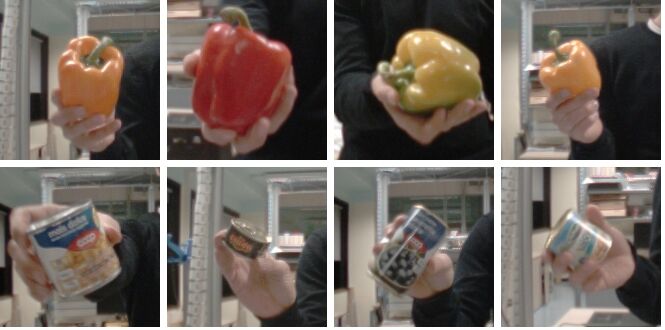}}
          \subfigure[CORe50]{
  \includegraphics[width=0.3\linewidth]{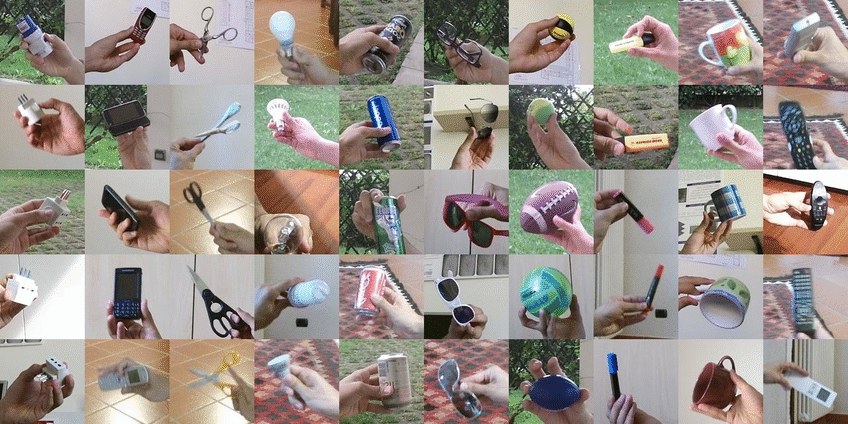}}
          \subfigure[CUB-200]{
  \includegraphics[width=0.3\linewidth]{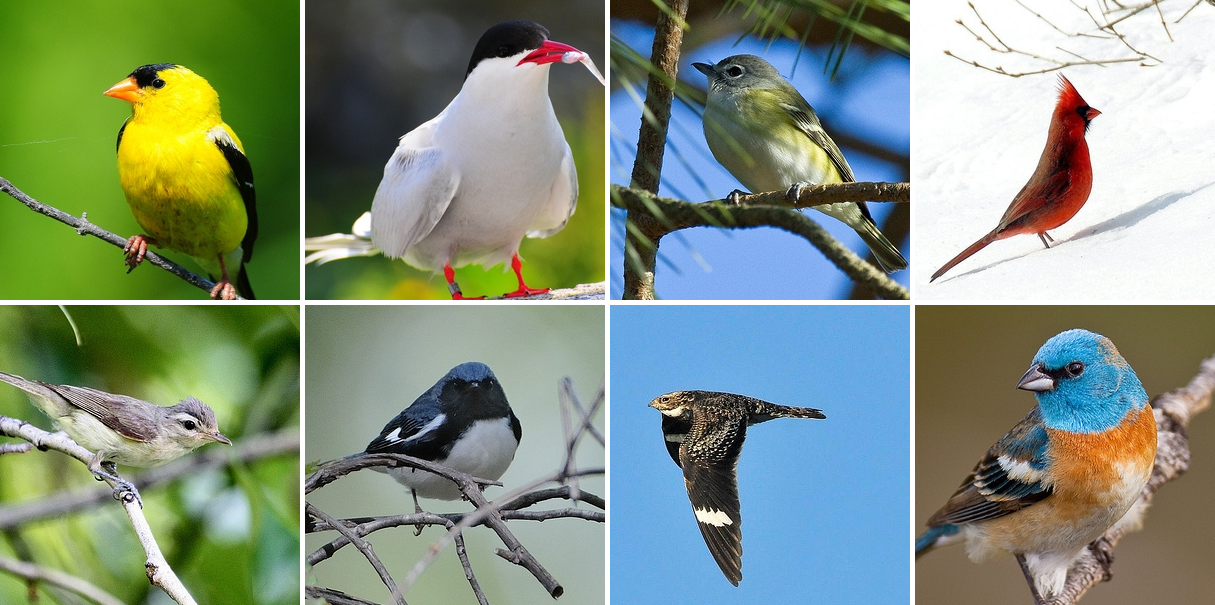}}
 
      \caption{Example images from each of the datasets used for evaluation.}
  \label{fig: datasets}
\end{figure*}

\section{Evaluation Protocol}
\label{sec:experiments}

We evaluate all methods in four different paradigms: 1) the data stream is randomly shuffled (iid), 2) the data stream is organized by class, 3) the data stream is temporally ordered by object instances (non-iid), and 4) the data stream is temporally ordered by object instances by class (non-iid). Paradigms 2 -- 4 will cause catastrophic forgetting in conventional DNNs~\cite{kemker2018forgetting}. For all paradigms, the model is required to learn sample-by-sample and is only allowed one pass through the entire training set. Every time it learns, we evaluate the model on all test data.

\subsection{Performance Metrics}
\label{subsec: metric} 

A streaming learner must be evaluated on its ability to learn quickly from possibly non-iid data streams. Kemker et al. recently introduced a metric for measuring the performance of incremental batch learners with respect to an offline baseline~\cite{kemker2018forgetting}, and we apply this metric to streaming learning here. Overall performance of a streaming learning method with buffer size $b$ is given by:
\begin{equation}
\Omega_{b} = \frac{1}{T} \sum_{t=1}^T \frac{\alpha_{t}}{\alpha_{\textrm{offline},t}} \enspace ,
\end{equation}
where $\alpha_{t}$ is the performance of the streaming classifier at time $t$, $\alpha_{\textrm{offline},t}$ is the performance of an optimized offline baseline at time $t$, and $T$ is the total number of testing events. An $\Omega_{b}$ value of 1 indicates that the streaming classifier performed as well as the offline model. It is possible for $\Omega_{b}$ to be greater than one, indicating that the streaming learner performed better than the offline model. $\Omega_{b}$ captures how well a model performs at various times during training, and because it is normalized, it is easier to compare performance across datasets/orderings. 

To evaluate each method's performance over a range of different buffer sizes, we use the metric given by:
\begin{equation}
\mu_{\textrm{total}} = \frac{1}{\left|\mathcal{B}\right|} \sum_{b\in \mathcal{B}} \Omega_{b}\enspace ,
\end{equation}
where $\mathcal{B}$ is the set of all buffer sizes tested. $\mu_{\textrm{total}}$ is an average over all buffer sizes tested. If $\mu_{\textrm{total}} = 1$, then a model performed as well as the offline model for all buffer sizes.

\subsection{Datasets}

We use two types of datasets for our experiments. The first type consists of two streaming learning datasets, iCub World 1.0~\cite{fanello2013icub} and CORe50~\cite{lomonaco2017core50}. These datasets are specifically designed for streaming learning and test the ability of each algorithm to learn from near real-time videos with temporal dependence. The second type is a fine-grain object recognition dataset with few training examples from 200 classes, i.e., CUB-200-2011~\cite{wah2011caltech}, which tests the ability of each algorithm to scale up to a large number of classes. Example images are provided in Fig.~\ref{fig: datasets}. For input features, we use embeddings from the ResNet-50~\cite{He_2016_CVPR} deep convolutional neural network (CNN) pre-trained on ImageNet-1K~\cite{russakovsky2015imagenet}. We use the 2048-dimensional features from the final mean pooling layer normalized to unit length.

\subsubsection{iCub World 1.0}

iCub1 is an object recognition dataset that contains household objects from ten different object categories~\cite{fanello2013icub}. Within each category, there are four particular object instances with roughly 200 images each, which are the frames from a video of a person moving the object around. iCub1 is ideal for streaming learning because it requires  learning from temporally ordered image sequences, which is naturally non-iid. In experiments, we use   $\mathcal{B}=\left\{2^{1},2^{2},\cdots , 2^{8}\right\}$. Overall, iCub1 has  10 classes, with 600-602 training images and 200-201 testing images per class.

\subsubsection{CORe50}

CORe50~\cite{lomonaco2017core50} is similar to iCub1, and both have 10 classes. However, CORe50 is more realistic since each class is made of 5 object instances and each is observed during 11 differing sessions (indoor, outdoor, various backgrounds, etc.). Each session contains a roughly 15 second video clip recorded at 20 fps. We use the cropped 128$\times$128 images and the train/test split suggested in \cite{lomonaco2017core50}, but sample the videos at 1 fps.
 The set of buffer sizes we use for each class is $\mathcal{B}=\left\{2^{1},2^{2},\cdots, 2^{8}\right\}$. This version of the dataset contains 10 classes, with 591-600 training images and 221-225 testing images per class.

\subsubsection{CUB-200-2011}

The Caltech-UCSD Birds-200 image classification dataset consists of 200 different species of birds with roughly 30 training images per class~\cite{wah2011caltech}. We use CUB-200 to examine how well models scale to a larger number of categories. We use  $\mathcal{B}=\left\{2^{1},2^{2},\cdots , 2^{4}\right\}$. 

\section{Results}
\label{sec: results}

\begin{table*}[t]
\caption{$\mu_{\textrm{total}}$ results. The top performer for each experiment is highlighted in \textbf{bold}. Values within one standard deviation of the top performer are highlighted in {\color{blue}{blue}}. Note that when ExStream was best, no methods were within one standard deviation. }
\begin{center}
\label{table: rehearsal results stream}
\bgroup
\setlength{\tabcolsep}{2pt}
\begin{adjustbox}{width=\linewidth,center}
\begin{tabular}{lccc >{\columncolor{Gray}}cccc>{\columncolor{Gray}}ccc>{\columncolor{Gray}}ccc>{\columncolor{Gray}}cc}
\toprule
\multirow{2}{*}{\textbf{Method}} & \multicolumn{4}{c}{\textbf{iid Ordered}} & \multicolumn{4}{c}{\textbf{Class iid Ordered}} & \multicolumn{3}{c}{\textbf{Instance Ordered}} & \multicolumn{3}{c}{\textbf{Class Instance Ordered}} & \textbf{Overall} \\
\cline{2-15} \\ [-0.9em]
 & \textbf{iCub1} & \textbf{CORe50} & \textbf{CUB-200} & \textbf{Mean} & \textbf{iCub1} & \textbf{CORe50} & \textbf{CUB-200} & \textbf{Mean} & \textbf{iCub1}\ & \textbf{CORe50} & \textbf{Mean} & \textbf{iCub1} & \textbf{CORe50} & \textbf{Mean} & \textbf{Mean} \\
\hline
    Reservoir Sampling & 0.902 & 0.857 & 0.690 & 0.816 & 0.898 & 0.809 & 0.696 & 0.801 & 0.932 & 0.891 & 0.912 & 0.902 & 0.799 & 0.851 & 0.838\\
    
    Queue & \textbf{0.960} & \textbf{0.959} & 0.703 & 0.874 & 0.866 & 0.760 & 0.702 & 0.776 & 0.810 & 0.911 & 0.861 & 0.733 & 0.677 & 0.705 & 0.808\\
    
    
    Online $k$-means & 0.914 & 0.893 & 0.768 & 0.858 & \textbf{0.955} & \textbf{0.927} & 0.769 & \textbf{0.884} & 0.967 & 0.903 & 0.935 & 0.915 & \textbf{0.892} & 0.904 & 0.890\\
    
    CluStream & 0.860 & 0.805 & 0.667 & 0.777 & 0.925 & 0.822 & 0.673 & 0.807 & 0.847 & 0.767 & 0.807 & 0.716 & 0.717 & 0.717 & 0.780\\
    
    HPStream ($\ell=1024$) & 0.916 & 0.874 & 0.754 & 0.848 & \textbf{0.955} & \color{blue}{0.916} & 0.753 & \color{blue}{0.875} & 0.960 & 0.877 & 0.919 & 0.883 & \color{blue}{0.879} & 0.881 & 0.877\\
    
    HPStream ($\ell=1536$) & 0.919 & 0.889 & 0.762 & 0.857 & \color{blue}{0.951} & \color{blue}{0.923} & 0.763 & \color{blue}{0.879} & 0.968 & 0.894 & 0.931 & 0.914 & \color{blue}{0.883} & 0.899 & 0.887\\
    
    ExStream & \color{blue}{0.953} & \color{blue}{0.951} & \textbf{0.789} & \textbf{0.898} & \color{blue}{0.953} & 0.868 & \textbf{0.790} & \color{blue}{0.870} & \textbf{0.989} & \textbf{0.950} & \textbf{0.970} & \textbf{0.969} & \color{blue}{0.882} & \textbf{0.926} & \textbf{0.909}\\

	\hline 
    
    No Buffer & 0.616 & 0.808 & 0.034 & 0.486 & 0.312 & 0.324 & 0.034 & 0.223 & 0.206 & 0.162 & 0.184 & 0.320 & 0.327 & 0.324 & 0.314\\
    
    Full Rehearsal & 0.977 & 0.984 & 0.951 & 0.971 & 1.004 & 1.001 & 0.955 & 0.987 & 1.006 & 1.033 & 1.020 & 1.001 & 1.011 & 1.006 & 0.992\\
    
    Offline & 1.000 & 1.000 & 1.000 & 1.000 & 1.000 & 1.000 & 1.000 & 1.000 & 1.000 & 1.000 & 1.000 & 1.000 & 1.000 & 1.000 & 1.000 \\
\bottomrule
\end{tabular}
\end{adjustbox}
\egroup
\end{center}
\end{table*}

Table~\ref{table: rehearsal results stream} shows the $\mu_{\textrm{total}}$ summary statistic for each model over all buffer sizes evaluated. Standard deviations were omitted due to space constraints, but they can be seen in the plots. For HPStream, $\ell$ represents the number of projected dimensions used in clustering. We chose projected dimensions that consisted of 50\% and 75\% of the 2048-dimensional features. The cluster structures used by CluStream and HPStream require \emph{twice} the amount of memory to maintain the same number of clusters as the other prototype methods. We ensured all algorithms stored the same number of clusters, independent of the total amount of memory required.  The offline DNN baseline performance is 79.47\%, 81.66\%, and 69.57\% for iid orderings of iCub1, CORe50, and CUB-200, respectively. See Appendix for more details. 

\subsection{Streaming iid Data}

\begin{figure*}[t]
\centering
\vspace{1.5mm}
\begin{tabular}
{c >{\centering\arraybackslash}m{1.5in} >{\centering\arraybackslash}m{1.5in} >{\centering\arraybackslash}m{1.5in} c}
	& iCub1 & CORe50 & CUB-200 & \\
    \rotatebox[origin=c]{90}{iid} &
	\includegraphics[scale=0.25]{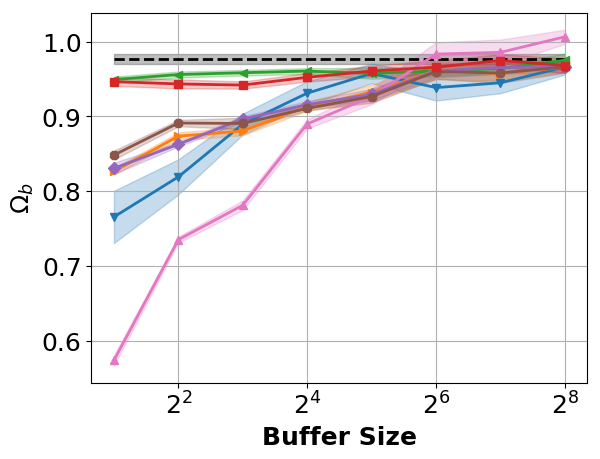} &
	\includegraphics[scale=0.25]{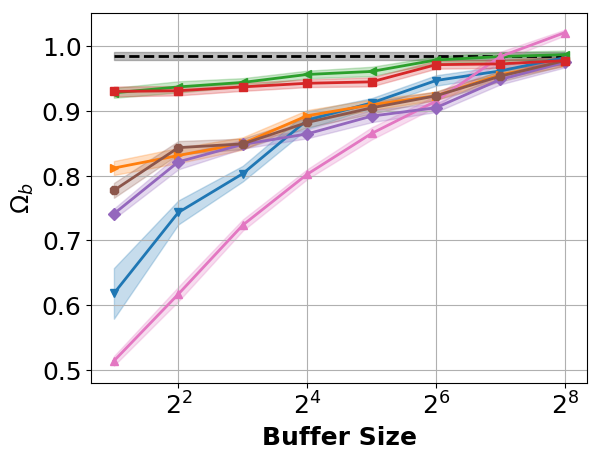} &
	\includegraphics[scale=0.25]{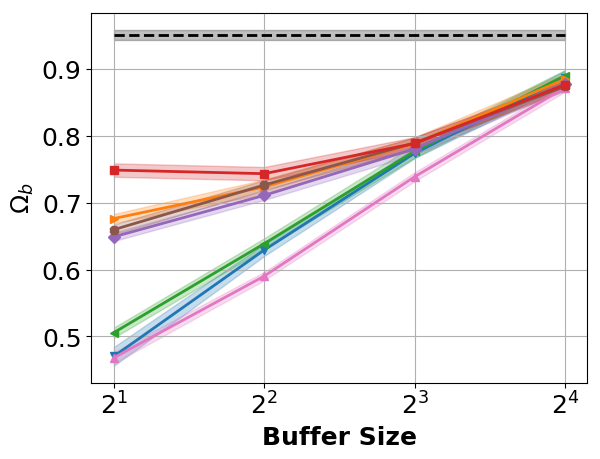} &
    \multirow{2}{*}{\includegraphics[scale=0.35]{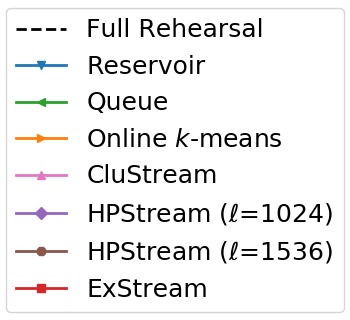}} \\
    \rotatebox[origin=c]{90}{class iid} &
	\includegraphics[scale=0.25]{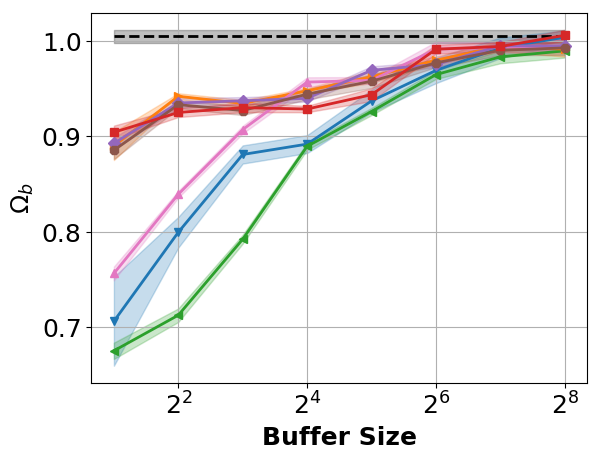} &
	\includegraphics[scale=0.25]{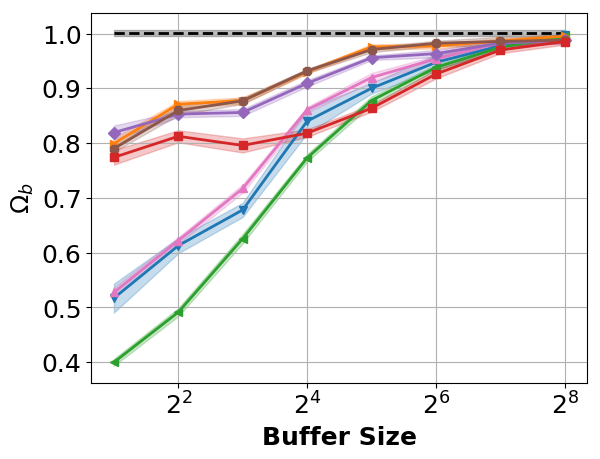} &
	\includegraphics[scale=0.25]{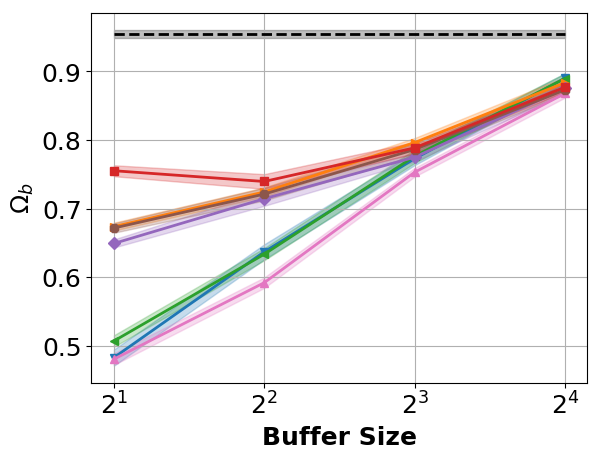} \\
\end{tabular}
\caption{Plots of $\Omega_{b}$ as a function of buffer size for the iid and class iid data orderings.}
\label{fig: streaming iid plots table}
\end{figure*}

The first experiment evaluates how well a streaming learner is able to learn quickly from a randomly shuffled data stream. Although this scenario is less realistic for a robotic learner, it resembles typical DNN training and should be easiest. See Fig.~\ref{fig: streaming iid plots table} for plots of $\Omega_{b}$ for this experiment.

Queue replacement performs best on the two streaming datasets, but ExStream performs best overall and on the harder CUB-200 dataset. We hypothesize the high performance of the queue model is because storing each sample as it comes in and then training a DNN on those iid samples is very similar to performing offline mini-batch stochastic gradient descent. With the exception of CluStream, which is not designed for high-dimensional data streams, all of the stream clustering methods yield significant performance advantages on the CUB-200 dataset, especially for small buffer sizes, demonstrating the efficacy of these algorithms. 

\subsection{Streaming Class iid Data}

The second experiment evaluates how well a learner is able to learn new classes over time. In this scenario, the data stream is organized by class, but the images for each class are randomly shuffled. Once an agent learns a class, it will never see that class again. This training paradigm is popular in incremental batch learning literature~\cite{li2016learning,lopez2017gradient,rebuffi2016icarl}, and will cause catastrophic forgetting in conventional DNNs. See Fig.~\ref{fig: streaming iid plots table} for $\Omega_{b}$ plots  for this experiment.

The best overall model was Online $k$-means, closely followed by ExStream and HPStream. ExStream again worked best on the harder CUB-200 dataset, performed well on iCub1, and also performed fairly well for small buffer sizes on CORe50. The plots in Fig.~\ref{fig: streaming iid plots table} demonstrate the advantage of Online $k$-means, HPStream, and ExStream over the other algorithms, especially for small buffer sizes.

\subsection{Streaming Instance Data}

\begin{figure}[t]
\centering
\begin{tabular}
{c >{\centering\arraybackslash}m{1.4in} >{\centering\arraybackslash}m{1.4in} c}
	& iCub1 & CORe50  & \\
    \rotatebox[origin=c]{90}{instance} &
	\includegraphics[scale=0.24]{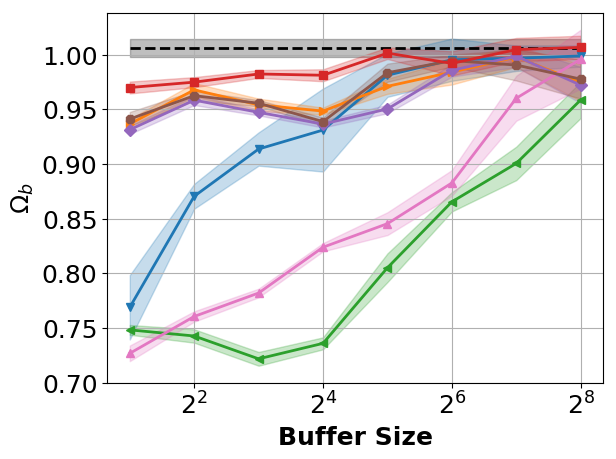} &
	\includegraphics[scale=0.24]{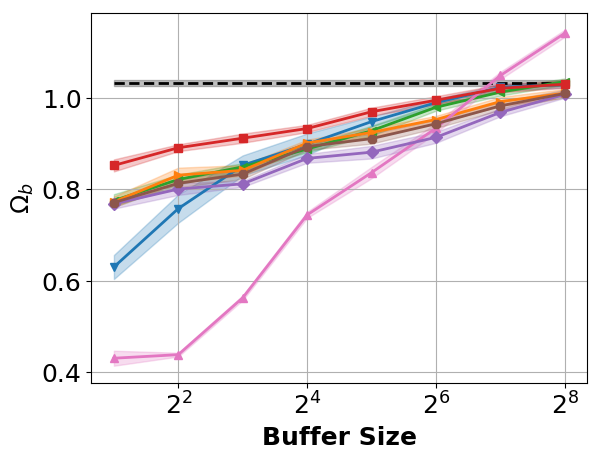} & \\
    \rotatebox[origin=c]{90}{class instance} &
	\includegraphics[scale=0.24]{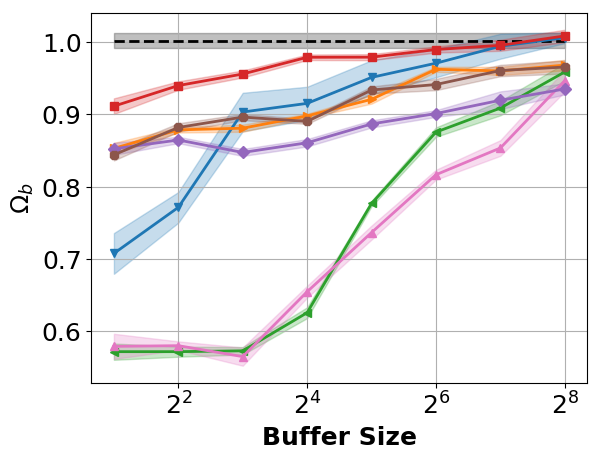} &
	\includegraphics[scale=0.24]{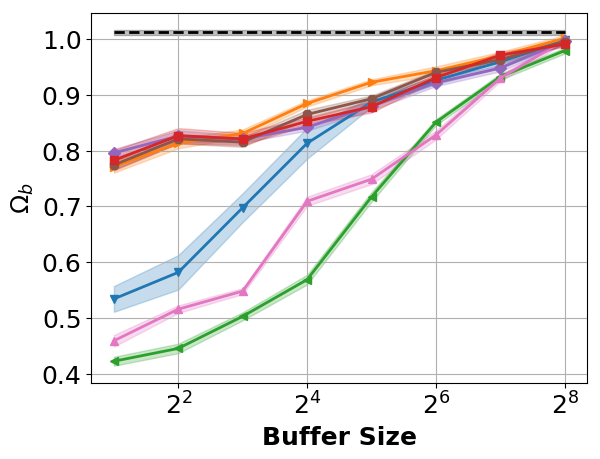} & \\
    & \multicolumn{2}{c}{\includegraphics[scale=0.3]{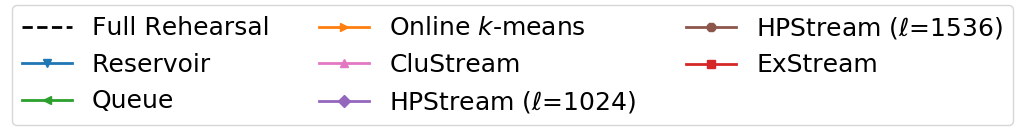}} \\
\end{tabular}
\caption{Plots of $\Omega_{b}$ as a function of buffer size for the instance and class instance data orderings.}
\label{fig: streaming instance plots table}
\end{figure}

The third experiment evaluates a learner in the most realistic robotic scenario. In this non-iid setting, the data stream is temporally ordered by object instances of different classes. While the agent cannot revisit previous object instances, objects from the same class may be observed at different times during training. That is, the agent may observe 50 images of dog \#1, then observe 10 images of cat \#3, and then observe 25 images of dog \#2, etc. This training paradigm will cause catastrophic forgetting in conventional DNNs. Fig.~\ref{fig: streaming instance plots table} contains plots of $\Omega_{b}$ versus buffer size for this experiment.

ExStream worked best overall. ExStream, Online $k$-means, and HPStream outperformed all other models on iCub1 for small buffer sizes. It is worth noting that CluStream performed better than full rehearsal for a buffer size of $2^8$ on CORe50. This result could indicate that CluStream needs larger buffer sizes. Since we are trying to explore memory efficient rehearsal, we are more interested in results where methods perform well with small buffer sizes.

\subsection{Streaming Class Instance Data}

The final experiment is a combination of the class and instance experiments where data are temporally ordered based on specific object instances within classes. That is, the robot would see all frames of all objects for class \#1, then all frames of all objects for class \#2, etc. In Fig.~\ref{fig: streaming instance plots table}, we provide plots of $\Omega_{b}$ as a function of buffer size for this experiment. ExStream performs best overall, although it performs slightly worse than Online $k$-means on CORe50.

\section{Discussion}
\label{sec:discussion}

We rigorously demonstrated that full rehearsal suffices for mitigating catastrophic forgetting in streaming learning paradigms designed to induce it when using high-resolution image datasets. For memory-limited rehearsal methods, we found that all were effective when the buffer was large and that there was more variance when the buffer was small. ExStream performed best across experiments, on average. ExStream requires half the memory of CluStream and HPStream, and it does not require any hyper-parameter tuning. Additionally, storing the new point rather than always merging it is particularly advantageous in the iid and several non-iid scenarios, where ExStream outperforms Online $k$-means. In the iid paradigm, queue worked surprisingly well, but it performed comparatively poorly for other orderings.

Stream clustering techniques designed for high-dimensional data performed best for non-iid data and maintained more consistent performance across both iid and non-iid data orderings. Stream clustering methods boosted performance on the more challenging CUB-200 dataset, especially with smaller buffer sizes. Performing better with smaller buffer sizes reduces the amount of memory and computational time necessary for learning with rehearsal, which is critical for learning on embedded devices. Future work should investigate methods for choosing \emph{subsets} of prototypes for training the DNN, as it is often slow to train with the entire buffer contents. Smart prototype selection strategies may improve performance by training the DNN on the most useful prototypes at a particular time-step.

We assumed a fixed buffer size per class, so adding more classes increases memory requirements. In the future, we plan to investigate using a single, fixed capacity, memory buffer shared among classes, enabling models to scale to larger datasets. Different classes could have differing numbers of prototypes, which may help with class imbalanced datasets. This approach could be extended to streaming regression tasks.   It would also be interesting to explore learning an exemplar storage policy for optimizing rehearsal.

\section{Conclusion}
\label{sec:conclusion}

Here, we demonstrated the effectiveness of rehearsal for mitigating catastrophic forgetting during streaming learning with DNNs. We also showed that rehearsal can be done in a memory efficient way by introducing the ExStream algorithm, and we demonstrated its efficacy on high-resolution datasets. ExStream is easy to implement, memory efficient, and works  well across iid and non-iid scenarios.

\FloatBarrier

\appendix
\label{sec: appendix}
\counterwithin{figure}{section}
\counterwithin{table}{section}
\counterwithin{subsection}{section}
\renewcommand{\thefigure}{A\arabic{figure}}
\renewcommand{\thetable}{A\arabic{table}}
\renewcommand{\thesubsection}{A.\arabic{subsection}}
\setcounter{figure}{0} 
\setcounter{table}{0} 

The parameters used to train each DNN for rehearsal are provided in Table~\ref{table: params}. All DNNs use batch norm. If the number of samples in the buffer is fewer than the batch size, we use a batch size of $\min{(\textrm{batch\_size},\textrm{num\_samples\_in\_buffer})}$.

We used the CluStream implementation from the stream~\cite{streamRpaper,streamRmanual} and streamMOA~\cite{streamMOAmanual} packages for R. For CluStream, we used the default parameter settings from \cite{streamMOAmanual}: horizon=1000 and maximal boundary=2. This implementation requires a $k$-means initialization of the clusters, which was done using a set of 2$\times$(buffer size) data points. For HPStream, we use the default parameter settings from \cite{aggarwal2003framework}: decay rate=0.5, spread radius factor=2, and speed=200. All other algorithms were implemented in Python 3.6.

\begin{table}[h]
\caption{Optimal parameters for each offline DNN.}
\begin{center}
\label{table: params}

\begin{tabular}{|l|c|c|c|}
\hline
\textbf{Parameter} & \textbf{iCub1} & \textbf{CORe50} & \textbf{CUB-200} \\
\hline\hline
Layer Sizes & [300, 150, 100] & [400, 100, 50] & [350, 300] \\
Batch Size & 256 & 256 & 100 \\
Weight Decay & 0.005 & None & None \\
Learning Rate & 0.0001 & 0.002 & 0.002 \\
Dropout & 0.5 & 0.5 & 0.75 \\
Activation & ReLU & ReLU & ELU \\
\hline
\end{tabular}
\end{center}
\end{table}

{\small
\bibliographystyle{ieee}
\bibliography{final_bibliography}
}

\end{document}